\documentclass[letterpaper, 10 pt, conference]{ieeeconf}  

\IEEEoverridecommandlockouts                              

\usepackage{graphics} 
\usepackage{svg} 
\usepackage{times} 
\usepackage{amsmath} 
\usepackage{amssymb}  
\usepackage{booktabs}
\let\labelindent\relax 
\usepackage{enumitem}
\usepackage{cuted}

\usepackage{mathtools}
\usepackage{tabularx}
\usepackage{array}
\usepackage{makecell}
\usepackage{multirow}
\usepackage{multicol}
\usepackage{graphicx}
    \graphicspath{{figure/}}
\usepackage[font={small}]{subcaption}
\usepackage[font={small}]{caption}
\usepackage{algorithm}
\usepackage{algpseudocode}
\IfFileExists{inconsolata.sty}{\usepackage[scaled=0.9]{inconsolata}}{} 
\usepackage{cite}

\makeatletter
\let\NAT@parse\undefined
\makeatother
\usepackage{url}
\usepackage{xcolor}
\definecolor{mycitecolor}{RGB}{71, 191, 38}
\definecolor{mylinkcolor}{RGB}{40, 115, 201}

\usepackage{hyperref}
\hypersetup{
  colorlinks=true,
  citecolor=mycitecolor,
  linkcolor=mylinkcolor,
  urlcolor=mycitecolor,
}

\usepackage[most]{tcolorbox}
\usepackage{listings}
\usepackage{xcolor}

\definecolor{codebg}{HTML}{F6F8FA}
\definecolor{codeframe}{HTML}{D0D7DE}
\definecolor{codekeyword}{HTML}{0000FF}
\definecolor{codecomment}{HTML}{008000}
\definecolor{codestring}{HTML}{A31515}

\newtcblisting{policycode}{
    enhanced,
    listing only,
    colback=codebg,
    colframe=codeframe,
    boxrule=0.6pt,
    arc=2pt,
    boxsep=2pt,
    left=3pt,
    right=3pt,
    top=0pt,
    bottom=0pt,
    listing options={
        language=Python,
        basicstyle=\ttfamily\scriptsize\color{black},
        keywordstyle=\color{codekeyword},
        commentstyle=\color{codecomment},
        stringstyle=\color{codestring},
        morekeywords={None,True,False},
        showstringspaces=false,
        breaklines=true,
        columns=fullflexible,
        keepspaces=true,
        tabsize=4,
        aboveskip=0pt,
        belowskip=0pt
    }
}

\title{\LARGE \bf
HuGo: LLMs as Whole-Body Policy Code Designers \\for Humanoid Loco-Manipulation
}

\author{Seoyeon Choi$^{1}$, Shizhao Ye$^{1,2}$, Nicholas Bui$^{1}$, Aayushi Shrivastava$^{1}$, Kanghyun Ryu$^{1}$, \\Dhruva Tirumala$^{3}$, Markus Wulfmeier$^{4}$, and Negar Mehr$^{1}$   
\thanks{$^{1}$University of California, Berkeley.
        {\tt\small \{seoyeon99, nicholasqbui, aayushis, kanghyun.ryu, negar\}@berkeley.edu}}%
\thanks{$^{2}$ShanghaiTech University.
        {\tt\small yeshzh2023@shanghaitech.edu.cn}}%
\thanks{$^{3}$Google DeepMind.
        {\tt\small dhruva92@gmail.com}}%
\thanks{$^{4}$Nomagic.
        {\tt\small mwulfmeier@nomagic.ai}. This work was conducted in part while the author was affiliated with Google DeepMind.}%
}

\begin{document}

\maketitle
\thispagestyle{empty}
\pagestyle{empty}

\preCutedStrip{\vskip-25pt}
\postCutedStrip{\vskip-8pt}
\begin{strip}
    \centering
    \hfill
    \begin{minipage}[t]{0.50\textwidth}
        \vspace{0pt}
        \centering
        \includegraphics[width=\linewidth]{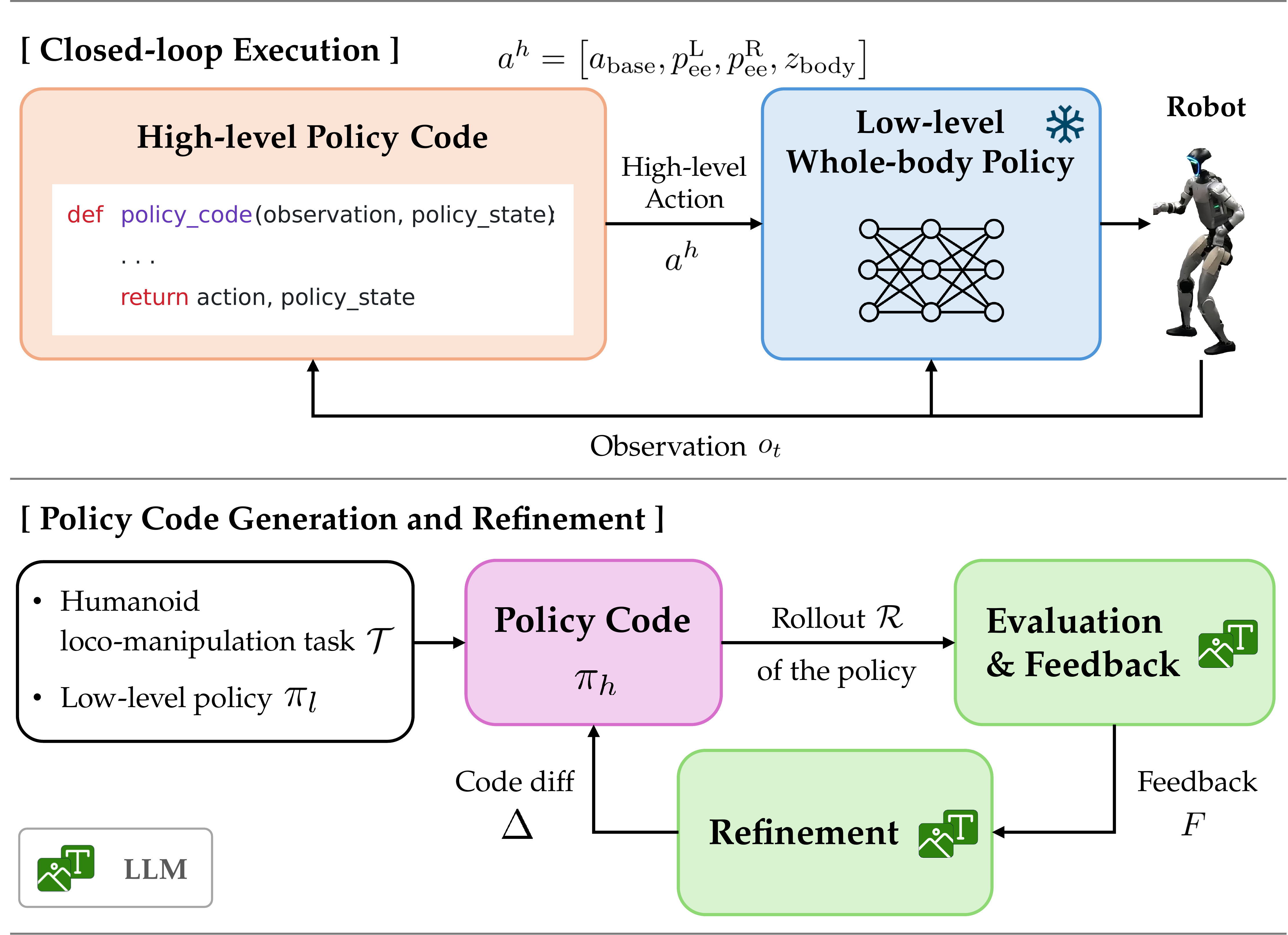}
    \end{minipage}
    \hfill
    \begin{minipage}[t]{0.46\textwidth}
        \vspace{0pt}%
        \centering
        \includegraphics[width=\linewidth]{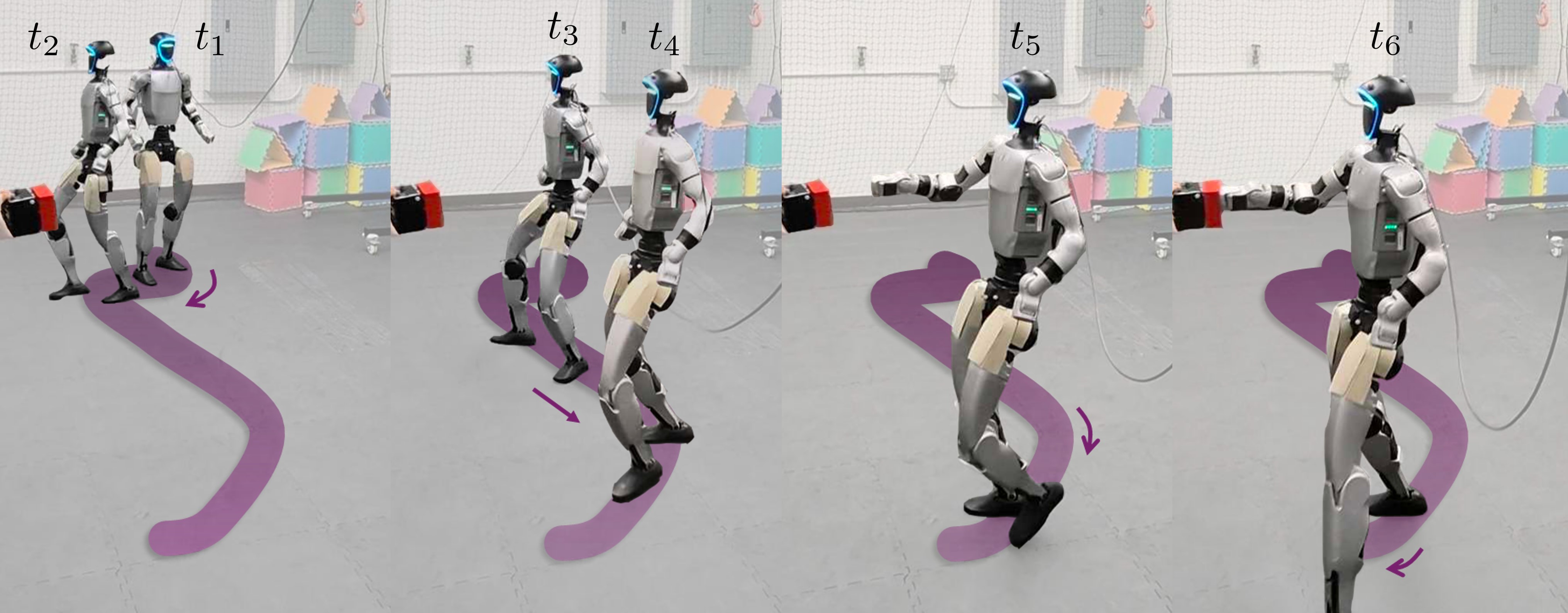}
        \includegraphics[width=\linewidth]{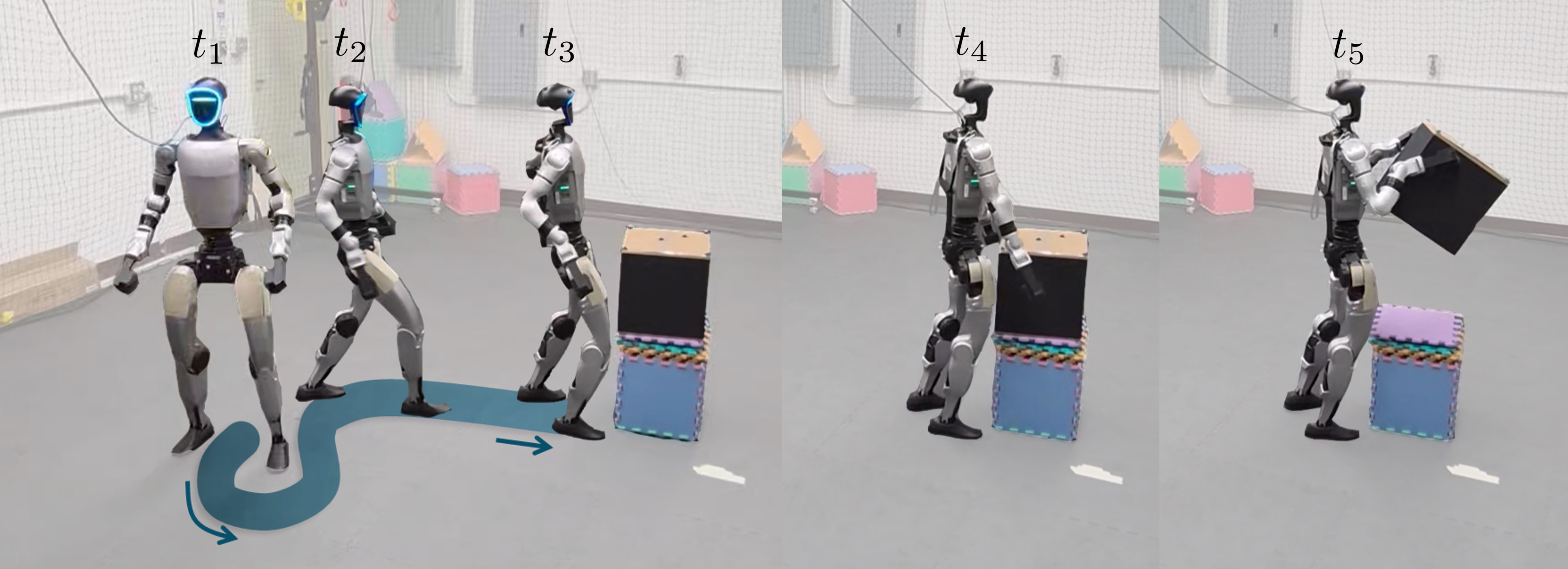}
    \end{minipage}

    \captionof{figure}{
    We introduce \textbf{HuGo}, a framework for generating closed-loop policy codes for humanoid loco-manipulation from task descriptions, without expert demonstrations, reference motions, or reward design. (Top left) The generated code outputs a high-level command $a^h$ at every timestep, consisting of planar locomotion, left and right end-effector positions, and body height, which a frozen low-level whole-body policy converts into joint actions. 
    (Bottom left) The LLM generates the policy code, diagnoses the policy's rollouts from numerical trajectories and selected video frames, and proposes a code diff, a localized update to the current policy rather than a full regeneration. (Right) Policies generated in simulation transferred zero-shot to hardware for \texttt{Push Button} (top) and \texttt{Lift Box} (bottom).
    } 
    \label{fig:main}
\end{strip}
\preCutedStrip{}
\postCutedStrip{}

\begin{abstract}

For humanoids to be useful in everyday environments, they must perform a wide range of tasks that couple locomotion and manipulation.
Existing approaches commonly acquire a loco-manipulation policy through reward engineering or demonstrations followed by task-specific training, making it costly to scale to new tasks.
In this work, we propose a hierarchical approach to humanoid loco-manipulation that eliminates these per-task requirements. HuGo, \underline{Hu}manoid policy code \underline{G}enerati\underline{o}n, uses a Large Language Model (LLM) to generate executable, closed-loop high-level policy code from a task description on top of a frozen low-level whole-body policy.
Given the task, observation, and command specifications, the LLM constructs the task logic in code. HuGo then refines the policy from its rollouts using numerical trajectories and selected video frames to produce feedback and targeted code updates.
Across five simulation tasks, using two different low-level policies, HuGo substantially outperforms a high-level reinforcement learning baseline and approaches the performance of a demonstration-based baseline. 
We achieve this level of performance without task-specific reward design or demonstration collection. We further demonstrate zero-shot transfer of simulation-generated policies to hardware and show that applying the same refinement loop to real-world rollouts can further improve transfer performance without expert demonstrations or policy retraining.
Project website is \href{https://iconlab.negarmehr.com/HuGo/} {https://iconlab.negarmehr.com/HuGo/}

\end{abstract}


\section{Introduction}

Recent advances in whole-body control have demonstrated what humanoid robots are capable of: they can now cartwheel, dance, and traverse rough terrain~\cite{liao2025beyondmimic,radosavovic2024terrain}. Yet some of the most useful things we would ask them to do are less spectacular: open a door, press an elevator button, or lift a box onto a shelf. These everyday tasks require a policy that coordinates locomotion and object interaction, deciding where to stand, how to reach, and when to move.

Obtaining such a policy remains a key challenge in humanoid loco-manipulation. Reinforcement learning (RL) optimizes policies against a reward function~\cite{zhao2025resmimic,kuang2025skillblender,li2026mpcrl,bohez2022imitate}, but can require substantial reward design~\cite{ma2024eureka} and repeated training to obtain the desired behavior. Demonstration-based methods guide learning with examples collected through teleoperation~\cite{seo2023deep} or extracted from human videos~\cite{weng2025hdmi, xie2026grail}, introducing the effort of collecting suitable data and, for human motions, retargeting them to the robot. Either way, the cost recurs for every task we want the humanoid to perform, as another reward to shape or another set of demonstrations to collect.

Recently, Code as Policies and subsequent work~\cite{liang2023code,fu2026capx,chen2026gap} have shown that Large Language Models (LLMs) can generate robot policy codes from natural-language instructions and produce manipulation policies without any training. 
We ask whether humanoid loco-manipulation can benefit in the same way. Directly controlling the joints of a high-dimensional robot such as a humanoid with an LLM is impractical~\cite{wang2023prompt,anthropic2026claude}. 
However, existing whole-body policies can already track locomotion, end-effector, and body height commands~\cite{yang2026handoff,luo2026sonic}. Our key insight is to have the LLM generate high-level policy code that outputs these commands, while reusing the frozen whole-body policy, avoiding motion data, reward engineering, and retraining for each new task.

To this end, we introduce \textbf{HuGo}, \underline{Hu}manoid policy code \underline{G}enerati\underline{o}n, which turns a task description into a closed-loop high-level policy \emph{without any task-specific training}. We prompt an LLM with the task description, the observations available to the robot, and the commands its low-level policy accepts, and have it generate policy code that reads the observation at every timestep and outputs the next command. \emph{Note that we provide the LLM with no example policies, no reference motions, and no library of task skills}, leaving the task logic entirely to the LLM. A policy generated from a task description is unlikely to be correct on the first attempt, so the LLM evaluates the execution of the policy from its numerical trajectory and selected video frames, and refines the code from that diagnosis.
This refinement loop takes as input nothing but executed trajectories, and can therefore work in both simulation and on real hardware.

We validate HuGo across five tasks and two low-level policies, in both simulation and on hardware. The generated policies substantially outperform a reinforcement learning baseline trained with hand-designed rewards, and approach the performance of a demonstration-based baseline on its own tasks. Policies generated in simulation can transfer zero-shot to hardware, and for a policy whose success drops after transfer, three refinement depths carried out directly on the physical robot raise it from $20\%$ to $90\%$.

In summary, our contributions are as follows:
\begin{enumerate}[leftmargin=8pt]
\vspace{-1pt}
    \item We propose HuGo, a framework in which an LLM generates closed-loop high-level policy code for humanoid loco-manipulation from a task description, on top of a frozen whole-body policy. This requires no expert demonstrations, reference motions, or task-specific reward design.
    \item We introduce a refinement loop that analyzes policy rollouts from numerical trajectories and selected video frames, and turns the resulting feedback into targeted code diffs.
    \item We validate HuGo across five loco-manipulation tasks and two low-level policies in simulation and on hardware, including zero-shot sim-to-real transfer and policy refinement performed directly from hardware rollouts.
\end{enumerate}

\section{Related Works}

\subsection{LLM-Guided Optimization}

As the reasoning ability of LLMs grows, recent work treats them as program optimizers guided by an evaluator~\cite{romera2024mathematical,ye2024reevo,novikov2025alphaevolve,karpathy2026autoresearch}. 
In robotics, LLMs have been used to generate rewards~\cite{ma2024eureka}, curricula~\cite{ryu2025curricullm}, or multi-agent coordination feedback~\cite{choi2025craft}, but still rely on costly task-specific RL training.
VLMs have also been used as the evaluator itself, judging whether a rollout succeeded from its video frames alone~\cite{du2023vlmsuccess}.
Following this line of work, we also use an LLM as an optimizer, but one that updates the policy code directly, guided by evaluations of the policy's rollouts.

\subsection{LLM-Generated Programmatic Robot Policies}

A more recent line of work uses the LLM to write the policy itself.
Code as Policies~\cite{liang2023code} first showed that an LLM can produce executable robot policies composing perception and control primitives. CaP-X~\cite{fu2026capx} benchmarks such LLMs on manipulation, where the model acts as a connection between specialized modules such as grasping and motion planning, and later work organizes these calls as a graph of skill nodes~\cite{chen2026gap} or improves the program across attempts by rollout fitness~\cite{sygkounas2026memento}. On humanoids, LLMs have produced behavior trees~\cite{wang2024autonomous} and sequences of locomotion and manipulation primitives~\cite{wen2025humanoidcoa}. 
In all of these, the generated program is assembled from a library of skills and modules supplied to the model.
We instead provide no primitives or modules to call. Given only descriptions of the policy observations and low-level policy's command interface, the LLM must decide where to stand, where to reach, and when to transition between stages. The resulting code runs as a closed-loop policy evaluated at every timestep.

\subsection{Humanoid Loco-Manipulation}

Humanoid loco-manipulation is commonly split into a low-level policy that tracks whole-body commands~\cite{luo2026sonic,yang2026handoff}, and a high-level policy that decides what to command. The high-level policy is often obtained with reinforcement learning, through residual learning over a pretrained motion-tracking policy~\cite{zhao2025resmimic}, blended pretrained skills~\cite{kuang2025skillblender}, or MPC-guided training~\cite{li2026mpcrl}. Others learn from human video priors~\cite{weng2025hdmi,xie2026grail}, collect demonstrations via MPC~\cite{schuck2026learning}, or plan contact sequences for trajectory optimization~\cite{taouil2026motiondisco}. Each new task, however, carries its own cost: a reward function, a demonstration and its retargeting, or a task-specific optimization problem. We instead obtain the high-level policy from a task description, on top of an unmodified low-level policy.

\section{Problem Formulation}

We consider a humanoid performing a loco-manipulation task $\mathcal{T}$ with a hierarchical structure: a high-level policy $\pi_h$ and a frozen low-level policy $\pi_l$. Our goal is to construct a task-specific $\pi_h$ that designs a closed-loop strategy for the task, and outputs commands that $\pi_l$ converts into joint-position targets.

We define a task as $ \mathcal{T} = \langle \ell_{\mathrm{task}}, \mathcal{X}, \rho, S \rangle $, where $\ell_{\mathrm{task}}$ describes the task in natural language, $\mathcal{X}$ is the state space of the humanoid, task objects, and the goal, $\rho$ is the distribution over initial states $x_0 \in \mathcal{X}$, which randomizes the initial robot, object, and goal poses, and $S$ is a binary success function indicating whether a rollout completes the task.

We assume that the high- and low-level policies are connected through a high-level command $a^h$ passed to $\pi_l$ at each timestep. We consider low-level policies that accept commands of the form:
\begin{equation}
\label{eq:command}
a^h
=
\left[
a_{\mathrm{base}},
p_{\mathrm{ee}}^{\mathrm{L}},
p_{\mathrm{ee}}^{\mathrm{R}},
z_{\mathrm{body}}
\right]
\in \mathbb{R}^{10},
\end{equation}
where $a_{\mathrm{base}}\in\mathbb{R}^{3}$ specifies planar locomotion and heading, $p_{\mathrm{ee}}^{\mathrm{L}}, p_{\mathrm{ee}}^{\mathrm{R}} \in \mathbb{R}^{3}$ indicate the desired positions of the left and right end-effectors, and $z_\mathrm{body} \in \mathbb{R}$ specifies the desired robot height. The command limits and coordinate frames are determined by $\pi_l$, and we provide them to $\pi_h$, which must produce commands consistent with them.

At each timestep $t$, the closed-loop high-level policy $\pi_h$ receives an observation $o_t$ and produces the high-level command $a_t^h$. The observation is
$o_t = \left[o_t^{\mathrm{robot}},o_t^{\mathrm{task}},a_{t-1}^{h}\right]$. 
Here, $o_t^{\mathrm{robot}}$ contains the base position and heading angle in the world frame, together with the measured base velocity, end-effector positions, and body height, while $o_t^{\mathrm{task}}$ contains task-relevant object and goal information, such as their poses, and $a_{t-1}^h$ is the previous high-level command.
The observation $o_t$ consists of numerical quantities rather than raw sensor data such as images. We assume it is available at each timestep, whether from simulation, external tracking, or vision.

We evaluate a high-level policy $\pi_h$ by executing it with the frozen low-level policy $\pi_l$ from an initial state $x_0\sim\rho$. Let $\mathcal{R}(\pi_h,\pi_l;x_0)$, written $\mathcal{R}$ for brevity, denote the resulting rollout. Our objective is to find the high-level policy $\pi_h$ that maximizes the expected task success: $ \pi_h^\star = \arg\max_{\pi_h}
\mathbb{E}_{x_0\sim\rho}
\left[
S\left(
\mathcal{R}(\pi_h,\pi_l;x_0)
\right)
\right].
\label{eq:objective} $

\section{Method}

To obtain a high-level policy $\pi_h$, we leverage the reasoning and task understanding of an LLM. However, an LLM cannot be queried at the rate a humanoid must be controlled, whereas the code it writes can run at that rate. Therefore, we represent $\pi_h$ as an \emph{executable policy code} that may maintain an internal state $m_t$ across timesteps:
\begin{equation}
\label{eq:highlevel_policy}
    \left(a^{h}_{t},m_{t}\right)
    =
    \pi_h\left(o_t,m_{t-1}\right).
\end{equation}
The internal state $m_t$ allows $\pi_h$ to retain task-relevant information across timesteps. Its structure is defined by the generated code and initialized at the beginning of each episode. In a Push Button task where the humanoid walks toward a button and presses it, for instance, $m_t$ may be a discrete phase that moves from approaching, to aligning, to pressing, or the commands issued at the previous timestep so that they can be smoothed.
In the following subsections, we describe in detail how our method generates $\pi_h$.


\subsection{Zero-shot Policy Generation} \label{subsec:zero_shot_policy}

For a given task $\mathcal{T}$, the generated policy code must specify a closed-loop strategy for the entire execution. In a Push Button task, the button appears at a different position, height, and facing direction in each episode, so a fixed sequence of commands may not be able to press it. The code must instead determine each command from the current observation, inferring an approach position from the button pose, deciding when to stop walking, and commanding the end-effector position and robot height needed to press it.

We provide the LLM with a natural-language prompt as $p =[\ell_{\mathrm{task}}, \ell_{\mathrm{robot}}, \ell_{\mathrm{obs}}, \ell_{\mathrm{act}}]$, where $\ell_{\mathrm{task}}$ describes the task, $\ell_{\mathrm{robot}}$ provides relevant robot information, such as its pelvis-to-head offset and arm length, and $\ell_{\mathrm{obs}}$ and $\ell_{\mathrm{act}}$ describe the observations available to $\pi_h$ and the high-level commands accepted by $\pi_l$, including their limits and coordinate frames. This prompt remains fixed throughout the pipeline. Importantly, \emph{it contains no demonstrations, example policy codes, action sequences, or reference implementations}. Given the prompt $p$, the LLM generates the zero-shot policy code $\pi_h^0$. The prompt alone, however, does not convey how these commands interact with $\pi_l$, the robot, and the environment, so policy logic that appears reasonable may still fail in execution due to tracking errors, contact dynamics, or poorly chosen transition conditions. We therefore examine how $\pi_h^0$ behaves through its rollouts, and diagnose the behavior.

\subsection{Rollout Evaluation and Feedback} \label{subsec:feedback}

We evaluate the current policy across $M$ initial states and measure its success rate using $S$ in simulation. This success rate shows how well the policy performs, but not why it succeeds or fails. Therefore, we record the rollouts and select the first unsuccessful rollout when it is available, since failures provide direct evidence about flaws in the policy logic. The selected rollout consists of $\mathcal{R} = (o_t,m_t,I_t)_{t=0}^{T}$, where $I_t$ is the video frame at timestep $t$ and $T$ is the termination timestep.

\textbf{Selecting what to show.}
The observation covers the robot's base and end-effector position, and the object poses, but not the full body configuration. Whether the humanoid leans off balance as it reaches for the button, for example, is invisible in the numbers but immediate in an image, so we provide the images to the LLM. A rollout spans thousands of timesteps, far more frames than can fit in context, and the relevant moments vary across tasks and policies. We therefore use the LLM to select task-relevant keyframes.

\textbf{Analyzing the rollout.}
Given the prompt $p$, the current policy code, and a sparse sample of $o_t$ and $m_t$ from $\mathcal{R}$, the LLM generates an executable analysis function $f_{\mathrm{analyze}}$. This function is generated specifically for the current rollout and policy, rather than fixed in advance. Applied to the complete trajectory, it returns a numerical summary $D$ and task-relevant keyframe indices $k_{\mathrm{key}}$. The summary is a string of at most 200 words containing quantities and events that the LLM determines may be useful for evaluating the rollout, such as distance extrema, threshold crossings, tracking errors, and internal-state transitions. Keyframes alone~\cite{hosseinzadeh2026kite} may omit the broader progression of the rollout, while evenly sampled frames~\cite{choi2025craft} may miss brief events such as a contact. We therefore combine both under a fixed frame budget $B$: when $f_{\mathrm{analyze}}$ returns fewer than $B$ keyframes, the remaining slots are filled with evenly spaced frames, giving the indices $k$.

\textbf{Generating feedback.}
The LLM receives the prompt $p$, the current policy code, the summary $D$, and the trajectory samples and video frames at $k$. From this, the LLM returns natural-language feedback $F$ describing what occurred during execution and the likely cause of success or failure. Note that this feedback rests on a single rollout and has no access to the scalar success rate of the current policy.

\subsection{Iterative Policy Code Refinement} \label{subsec:refinement}

Given the rollout feedback $F$, rather than regenerating the entire policy, we have the LLM propose a localized code diff $\Delta$ conditioned on the current policy code. This preserves policy logic that works while modifying only the parts that $F$ identifies as a fault, and keeps the generated output concise.

Such a modification is not guaranteed to improve the policy, so refinement proceeds over multiple attempts. Note that failed attempts are still informative. Providing the LLM with previously attempted modifications and their evaluations helps it avoid repeating them. However, accumulating every attempt across the entire process would cause the context to grow continually. We therefore keep this history of failed attempts only while it remains relevant, that is, while the LLM is still trying to improve the same policy.

We organize refinement into \emph{depths} and \emph{rounds}. A depth is one improvement over the policy it starts from, and it is given up to three rounds to find one. In each round, the LLM proposes a diff, and the resulting candidate is evaluated as in Sec.~\ref{subsec:feedback}. If the candidate reaches a higher success rate, it is accepted: it becomes the current policy, the history is cleared, and the next depth begins from it. If not, its diff, success rate, and feedback are added to the history, and the next round proposes a different modification to the same policy.

Refinement terminates when either the policy reaches $100\%$ success, or when the maximum refinement depth $D_{\max}$ is reached, or when all three rounds in one depth fail to improve the current policy.
We stop after three failed rounds because the task logic a policy code encodes cannot always be repaired by incremental changes. In all cases, we return the current best-performing policy as the final refined high-level policy. Algorithm~\ref{alg:method} summarizes the complete procedure, from zero-shot generation through refinement, that yields a single high-level policy.

\begin{algorithm}[t]
\caption{Generation of a Single High-Level Policy}
\footnotesize
\label{alg:method}
\begin{algorithmic}[1]
\Require Prompt $p$, frozen low-level policy $\pi_l$, $M$ initial states,
         maximum refinement depth $D_{\max}$
\Ensure Refined high-level policy $\pi_h$

\State Generate zero-shot policy $\pi_h$ from $p$
\State Evaluate $\pi_h$ over $M$ initializations to obtain success rate $s$

\For{$d = 1,\ldots,D_{\max}$}
    \State \textbf{if} $s = 100\%$ \textbf{then return} $\pi_h$

    \State Diagnose a rollout of $\pi_h$ as in Sec.~\ref{subsec:feedback} to obtain feedback $F$
    \State Initialize the refinement context with $p$, $\pi_h$, and $F$

    \For{$r = 1,2,3$}
        \State Generate code diff $\Delta$ from the refinement context
        \State Apply $\Delta$ to $\pi_h$ to obtain candidate $\tilde{\pi}_h$
        \State Evaluate $\tilde{\pi}_h$ over $M$ initializations
               to obtain success rate $\tilde{s}$

        \If{$\tilde{s} > s$}
            \State $\pi_h \gets \tilde{\pi}_h$, $s \gets \tilde{s}$
            \State \textbf{break}
        \ElsIf{$r = 3$}
            \State \Return $\pi_h$
        \Else
            \State Diagnose a rollout of $\tilde{\pi}_h$ to obtain feedback $\tilde{F}$
            \State Add $\Delta$, $\tilde{s}$, and $\tilde{F}$ to the refinement context
        \EndIf
    \EndFor
\EndFor

\State \Return $\pi_h$

\end{algorithmic}
\end{algorithm}

\subsection{Best-of-$N$ Selection}

LLM generation is stochastic, so independent runs produce policies with qualitatively different task logic. Refinement edits a policy locally and rarely replaces the logic it starts from, so the final performance depends on that initial logic. Therefore, we run the full pipeline $N$ times, each from a new zero-shot policy, and keep the one with the highest success rate. Even repeated $N$ times, the procedure requires no human intervention and costs on the order of ten dollars and a few hours, as we show in Sec.~\ref{subsec:cost}.

\begin{figure*}
    \centering
    \begin{subfigure}[b]{0.19\linewidth}
        \centering
        \includegraphics[width=\linewidth]{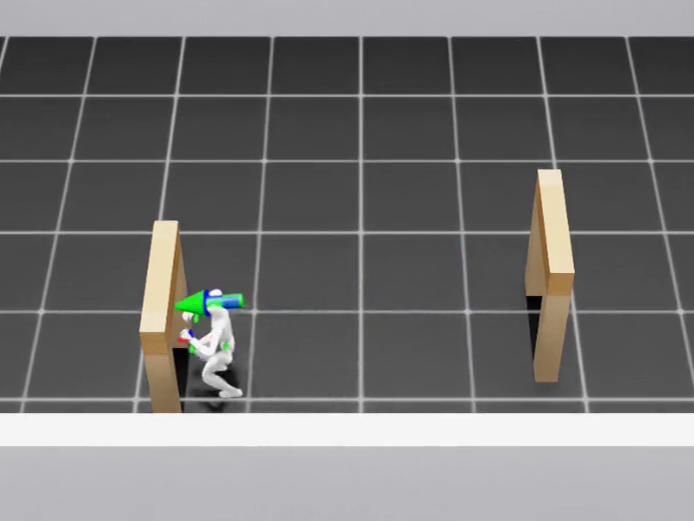}
        \caption{Low Gate Passing}
    \end{subfigure}
    \hfill
    \begin{subfigure}[b]{0.19\linewidth}
        \centering
        \includegraphics[width=\linewidth]{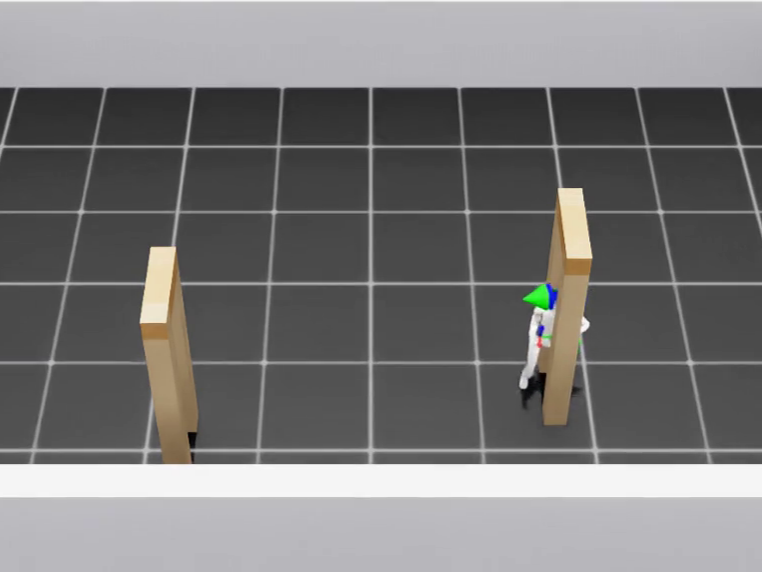}
        \caption{Narrow Gate Passing}
    \end{subfigure}
    \hfill
    \begin{subfigure}[b]{0.19\linewidth}
        \centering
        \includegraphics[width=\linewidth]{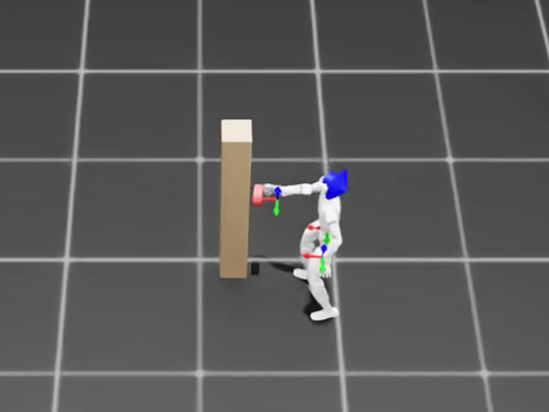}
        \caption{Push Button}
    \end{subfigure}
    \hfill
    \begin{subfigure}[b]{0.19\linewidth}
        \centering
        \includegraphics[width=\linewidth]{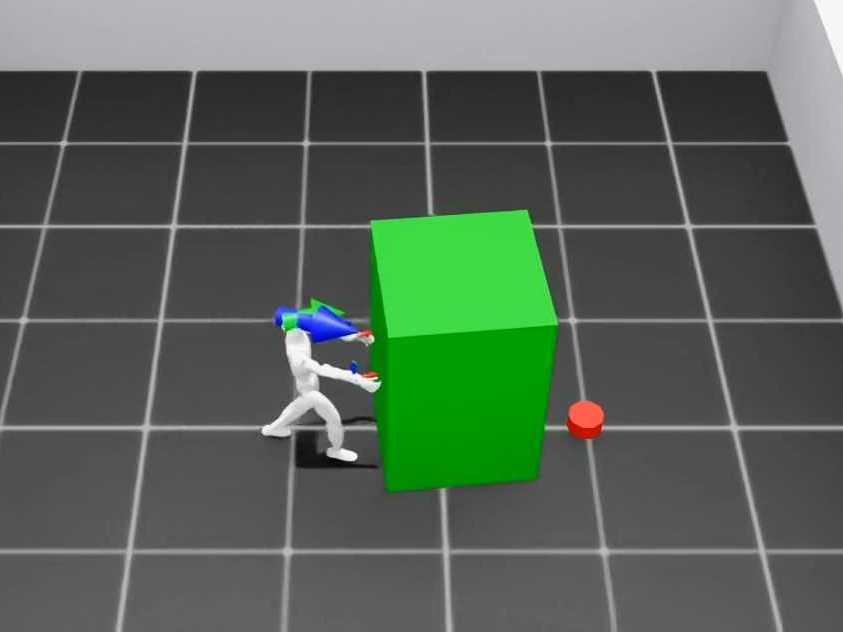}
        \caption{Push Box}
    \end{subfigure}
    \hfill
    \begin{subfigure}[b]{0.19\linewidth}
        \centering
        \includegraphics[width=\linewidth]{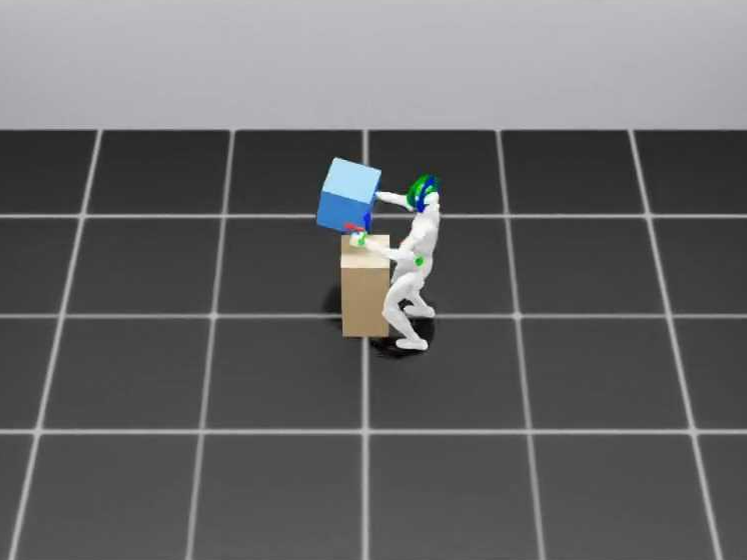}
        \caption{Lift Box}
    \end{subfigure}
    \caption{The five simulated loco-manipulation tasks. All are situated in an indoor environment and randomize their initial conditions, so a policy must adapt its navigation path, approach pose, and command timing to each episode.}
    \label{fig:five_task}
\end{figure*}

\section{Simulation Experiments}

\begin{table*}[t]
    \centering
    \scriptsize
    \begin{tabular}{ll|c|c|c|c|c}
        \toprule
        Low-level & Method & \texttt{Low Gate Passing} & \texttt{Narrow Gate Passing} & \texttt{Push Button} & \texttt{Push Box} & \texttt{Lift Box} \\
        \midrule
        \multirow{3}{*}{\texttt{rl\_isaac}} & High-level RL &  $11.33 \pm 1.70$ & $0.0 \pm 0.0$ & $7.33 \pm 1.70$ & $1.67 \pm 1.25$ & $0.0 \pm 0.0$ \\
         & HuGo (no refinement) & $100.00 \pm 0.00$ & $67.15 \pm 14.13$ & $73.92 \pm 7.56$ & $49.31 \pm 17.49$ & $44.62 \pm 25.22$ \\
         & \textbf{HuGo} & $\boldsymbol{100.00 \pm 0.00}$ & $\boldsymbol{96.20 \pm 1.31}$ & $\boldsymbol{79.97 \pm 1.68}$ & $\boldsymbol{89.33 \pm 11.37}$ & $\boldsymbol{75.92 \pm 16.44}$\\
        \midrule
        \multirow{2}{*}{\texttt{sonic}} & HuGo (no refinement) & - & $65.93 \pm 21.98$ & $48.70 \pm 16.43$ & $46.30 \pm 25.86$ & $11.00 \pm 3.79$ \\
         & \textbf{HuGo} & - & $\boldsymbol{97.54 \pm 1.97}$ & $\boldsymbol{85.78 \pm 7.89}$ & $\boldsymbol{61.79 \pm 12.18}$ & $\boldsymbol{24.91 \pm 6.12}$ \\
        \bottomrule
    \end{tabular}
    \caption{Success rates (\%) on the five-task suite with best-of-$10$ selection. Reported success rates are evaluated on $100$ held-out initial states not used for policy refinement or best-of-N selection. The same held-out initial states are used for all methods. HuGo (no refinement) selects among the same policies before refinement. \texttt{sonic} is not evaluated on \texttt{LGP} as it cannot change its height while walking.}
    \label{tab:full_SR}
    \vspace{-5pt}
\end{table*}

\subsection{Experiment Setup}

We evaluate our method on five humanoid loco-manipulation tasks shown in Fig.~\ref{fig:five_task}. These tasks are situated in an indoor environment where the surrounding walls constrain the available paths and require the policy to account for obstacles during navigation: 
\begin{itemize}[leftmargin=8pt]
    \item \texttt{Low Gate Passing (LGP)}: The humanoid must pass through two low gates by lowering its body height.
    \item \texttt{Narrow Gate Passing (NGP)}: The humanoid must pass through two narrow gates by turning its body sideways.
    \item \texttt{Push Button (PBT)}: The humanoid must approach a button and press it using one of its end-effectors.
    \item \texttt{Push Box (PBX)}: The humanoid must approach a box and push it toward a target within $10\,\mathrm{cm}$ and hold it there for one second.
    \item \texttt{Lift Box (LBX)}: The humanoid must approach a box resting on a podium, grasp it, and lift it above $1.0\,\mathrm{m}$.
\end{itemize}
In every task we randomize the initial poses of the humanoid, objects, and goals. This changes the navigation path, approach pose, and timing of the required commands, requiring the policy to adapt accordingly. These tasks also pose distinct challenges: the gate-passing tasks emphasize navigation and whole-body configuration through constrained spaces, whereas the remaining tasks require the policy to coordinate navigation with precise object interaction.

We evaluate our method using two frozen low-level policies with different command specifications, demonstrating our method's compatibility with different low-level policies:
\begin{itemize}[leftmargin=8pt]
    \item \textbf{RL-based policy} (\texttt{rl\_isaac}): We train a low-level policy  using dense rewards for tracking the high-level commands. The base command specifies planar linear and yaw velocities, the end-effector targets are expressed in the torso frame, and the height specifies the desired torso height. The reward design is adapted from low-level policy in ~\cite{liu2025ego}.
    \item \textbf{SONIC 3-Point VR} (\texttt{sonic}): We use the 3-point VR controller from SONIC~\cite{luo2026sonic}, a state-of-the-art motion-tracking policy trained on a large-scale human-motion dataset. The base command specifies planar linear velocity and a yaw heading, the end-effector targets are expressed in the pelvis frame, and the height specifies the desired pelvis height. Since SONIC cannot change its height while walking, we do not evaluate it on \texttt{LGP}.
\end{itemize}

\textbf{Baselines.} We compare against two alternative ways of obtaining a high-level policy. \textbf{High-level RL}: a high-level policy trained with PPO~\cite{schulman2017proximal} for 384M environment steps, using the same observations, command space, and low-level policy as ours, with a hand-designed dense reward for each task. \textbf{HDMI}~\cite{weng2025hdmi}: a demonstration-based method that learns interactive humanoid whole-body control by imitating reference motions extracted from human videos. We evaluate against HDMI on two of its tasks that differ in where the difficulty lies, one in navigation and other in manipulation:
\begin{itemize}[leftmargin=8pt]
    \item \texttt{Push Door with Hand (PDH)}: The humanoid must approach a door, push it open, and walk through.
    \item \texttt{Carry and Place Bread Box (CPB)}: The humanoid must grasp a bread box, carry it, and place it at a target.
\end{itemize}
These tasks fix the initial pose of the humanoid relative to the object.
For \texttt{CPB}, we reduce the bread box mass to a payload value supported by our low-level policies.

\textbf{Implementation details.} All experiments use a Unitree G1 in IsaacSim on a single NVIDIA L40S GPU. The generated policy code runs at $50\,\mathrm{Hz}$ with \texttt{rl\_isaac} and at $10\,\mathrm{Hz}$ with \texttt{sonic}, matching each deployment stack. We use \texttt{gpt-5.4} for the LLM, and each policy is evaluated on a fixed set of $M=100$ initial states. The reported final results use a separate held-out set. The evaluator receives $B=10$ frames per rollout. Refinement runs for at most $D_{\max}=7$ depths with three rounds each. For each task, we independently generate and refine $30$ policies, and report best-of-$10$ results by averaging over $2000$ random subsets of $10$ of these policies.
\begin{table}[t]
    \centering
    \scriptsize
    \begin{tabular}{l|c|c}
        \toprule
         &  \texttt{PDH} & \texttt{CPB}\\
         \midrule
         HDMI~\cite{weng2025hdmi} & $99.95 \pm 2.21$ & $98.4 \pm 12.6$ \\
         \midrule
         HuGo & $100.0 \pm 0.00$ & $81.25 \pm 26.23$ \\
         \bottomrule
    \end{tabular}
    \caption{Success rates (\%) on two HDMI tasks. HDMI trains a policy per task from a human demonstration, while our policies are generated from a task description without demonstrations. HuGo is reported with best-of-$10$ selection.}
    \label{tab:hdmi_tasks}
\end{table}

\begin{figure*}[t]
    \centering
    \newlength{\plotpanelh}\setlength{\plotpanelh}{1.42in}%
    \begin{minipage}[b]{0.345\textwidth}
        \centering
        \includegraphics[height=\plotpanelh]{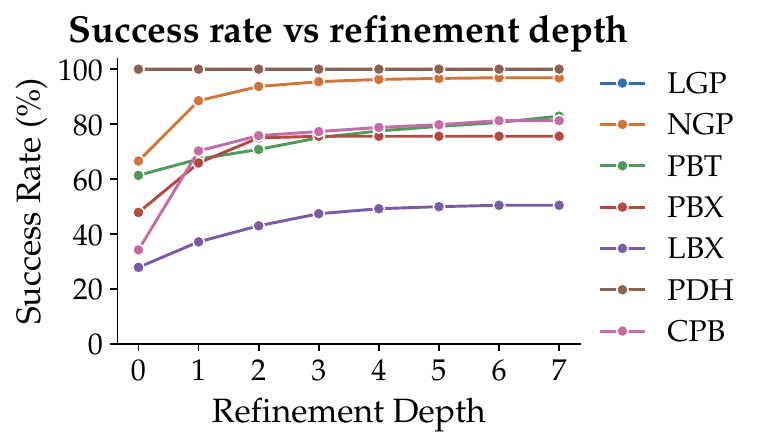}
        \captionof{figure}{Best-of-$10$ success rate versus refinement depth, averaged over low-level policies. Depth $0$ corresponds to the policies before refinement.}
        \label{fig:depth}
    \end{minipage}
    \hfill
    \begin{minipage}[b]{0.63\textwidth}
        \centering
        \begin{subfigure}[b]{0.49\linewidth}
            \centering
            \includegraphics[height=\plotpanelh]{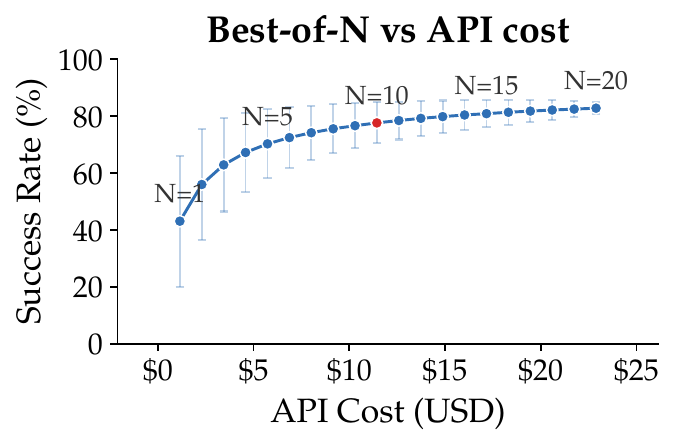}
        \end{subfigure}
        \hfill
        \begin{subfigure}[b]{0.49\linewidth}
            \centering
            \includegraphics[height=\plotpanelh]{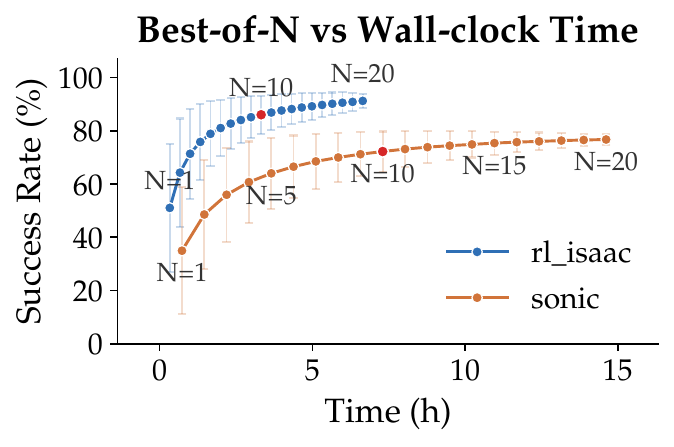}
        \end{subfigure}
        \caption{Best-of-$N$ success rate against the cost of obtaining it. Left: API cost, averaged over the five tasks and both low-level policies, where error bars denote the standard deviation across runs and the red marker indicates the operating point $N=10$. Right: wall-clock time, shown separately for each low-level policy.}
        \label{fig:cost}
    \end{minipage}
\end{figure*}

\begin{figure*}[t]
    \centering
    \begin{minipage}[b]{0.625\textwidth}
        \centering
        \scriptsize
        \begin{tabular}{l|c|c|c|c|c}
            \toprule
             &  \texttt{NGP} & \texttt{PBT} & \texttt{PBX} & \texttt{LBX} & Avg. \\
             \midrule
             Zero-shot & $0.0$ & $1.0$ & $0.0$ & $0.0$ & $0.25$ \\
             \midrule
             Text Only & $47.0 \pm 38.8$ & $53.2 \pm 23.9$ & $19.6 \pm 36.7$ & $1.4 \pm 2.3$ & $30.3$\\
             Even Only & $52.4 \pm 42.9$ & $55.4 \pm 29.3$ & $1.8 \pm 1.3$ & $13.0 \pm 18.2$ & $30.6$ \\
             Keyframe Only & $56.6 \pm 46.3$ & $45.6 \pm 28.3$ & $5.4 \pm 6.2$ & $20.2 \pm 16.0$ & $32.0$\\
             \midrule
             Ours & $\boldsymbol{56.8 \pm 46.4}$ & $\boldsymbol{75.8 \pm 4.6}$ & $\boldsymbol{42.4 \pm 34.1}$ & $\boldsymbol{35.6 \pm 31.9}$ & $\boldsymbol{52.6}$ \\
             \bottomrule
        \end{tabular}
        \captionof{table}{Success rates (\%) after refinement when the rollout evaluation is given different information, mean $\pm$ standard deviation over five runs per task using \texttt{rl\_isaac}. All variants start from the same zero-shot policies, shown in the first row.}
        \label{tab:ablation_evaluation}
    \end{minipage}
    \hfill
    \begin{minipage}[b]{0.35\textwidth}
        \centering
        \newlength{\pbtpanelh}\setlength{\pbtpanelh}{0.85in}%
        \includegraphics[height=\pbtpanelh]{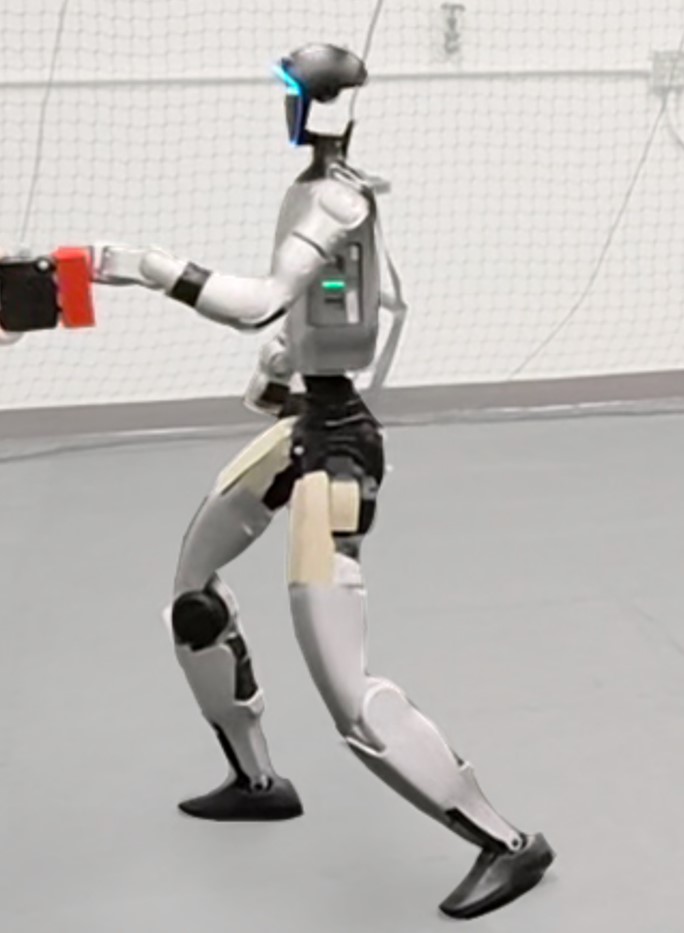}\hfill
        \includegraphics[height=\pbtpanelh]{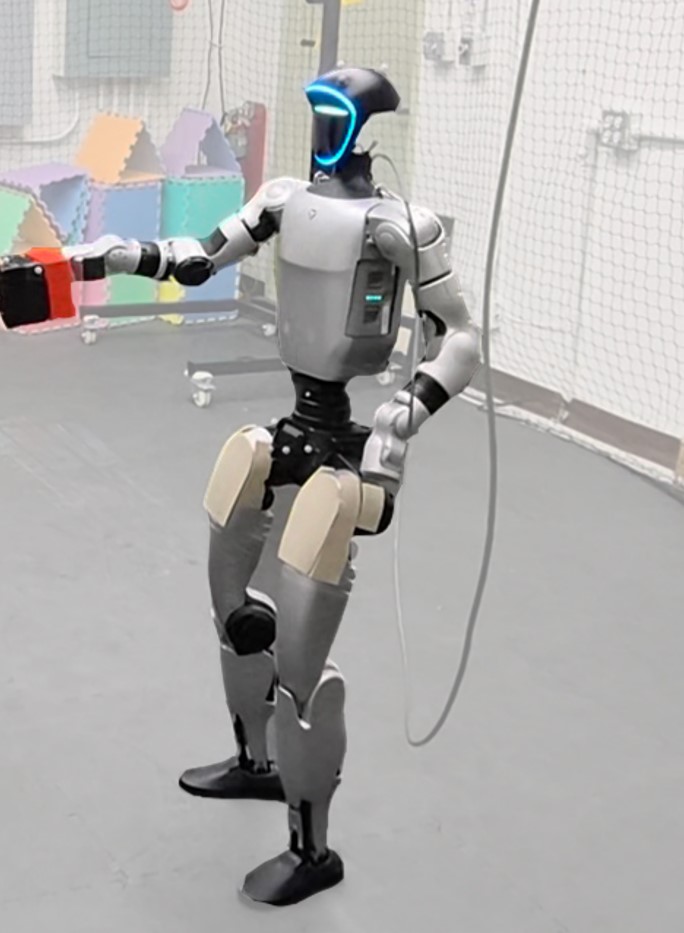}\hfill
        \includegraphics[height=\pbtpanelh]{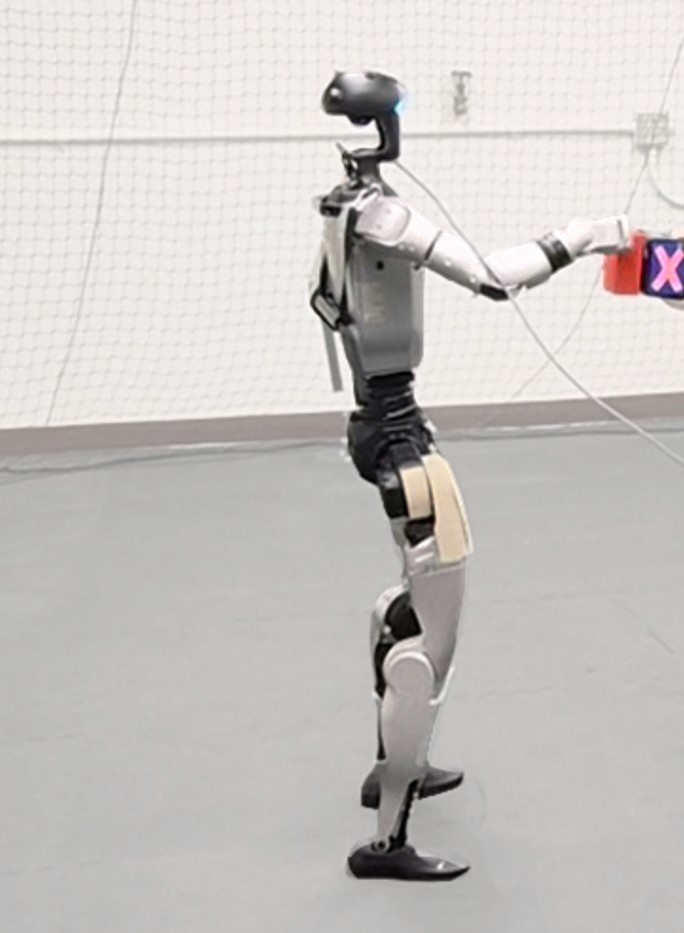}%
        \caption{\texttt{PBT} on hardware, all from the same policy code. The button position and facing direction determine which side the policy approaches from and which end-effector it presses with.}
        \label{fig:pbt_conditions}
    \end{minipage}
    \vspace{-10pt}
\end{figure*}

\subsection{Main Results} \label{subsec:main_results}

We report success rates on our five-task suite in Table~\ref{tab:full_SR}. With the \texttt{rl\_isaac} low-level policy, our method achieves near or above $80\%$ success on four of the five tasks, including near-perfect success on both gate-passing tasks. Our method also works with \texttt{sonic} without any change to the pipeline since switching low-level policies only requires describing the new command interface in the prompt. \texttt{LBX} remains the most challenging task, and the hardest with \texttt{sonic}, for reasons we examine in Remark.
\emph{Note that all of these policies are obtained from a task description, without a single expert demonstration, motion reference, or reward function.}

Refinement is essential to these results. Compared with selecting among policies before refinement, refinement improves the success rate by up to $40$ percentage points. Refinement also substantially reduces the standard deviation in Table~\ref{tab:full_SR} for most of the tasks as it raises the success rate of policies across runs rather than of a few outliers.
Fig.~\ref{fig:depth} shows how this improvement accumulates over refinement depths: most of the gain occurs within the first two depths, while harder tasks such as \texttt{LBX} continue to improve until the final depth.


\textbf{Remark}. LBX exposes a limitation in the LLM's diagnosis of contact failures. The generated code often moves the end-effectors directly to the grasp width without first establishing clearance around the box, causing them to sweep against its sides. \texttt{rl\_isaac} tracks accurately enough that they only brush the box, while \texttt{sonic}, with $4\,\mathrm{cm}$ tracking error, pushes it off the podium. Although the LLM identifies the box being pushed off in evaluation, its feedback does not attribute this to the arm approach, so refinement never targets it. Adding a short geometry hint, without changing the interface or the low-level policy, raises the best-of-$10$ success rate with \texttt{sonic} from $24.9\%$ to $72.3\%$, which is comparible to the \texttt{rl\_isaac} results. The existing policy interface can therefore express an effective correction, but the LLM does not reliably infer it from rollout feedback without task-specific guidance. Another limitation comes from both low-level policies, which struggle to maintain the contact forces needed to hold the box, so it slips out during lifting.

\textbf{Comparison with High-Level RL.} Table~\ref{tab:full_SR} reports the high-level RL baseline over three training seeds. The trained policies remain at low success rates, far below the policies our method generates. Their failures are informative. \texttt{PBX}, for instance, requires the box to stay on the target for a full second, and the trained policy does bring the box to the target but commonly overshoots and pushes it straight off again. Across tasks, reinforcement learning discovers how to act on the object but not the staged structure the task requires, approach, align, act, and hold, which is exactly what our generated policy code specifies explicitly.

\textbf{Comparison with Demonstration-Based Methods.} Demonstration-based methods are a popular way to obtain humanoid loco-manipulation skills. We ask whether HuGo, which receives a task description, can approach the performance of methods that are directly trained to mimic a given demonstration. Table~\ref{tab:hdmi_tasks} reports the results on \texttt{PDH} and \texttt{CPB}. On \texttt{PDH}, our method matches HDMI, with both succeeding in essentially all episodes. 
On \texttt{CPB}, our method reaches $81.25\%$ from a task description without demonstrations, against $98.4\%$ for HDMI, which imitates a single retargeted human demonstration of the task.
The failures on \texttt{CPB} come from the contact-force limitation described above. \texttt{rl\_isaac} tracks the end-effectors accurately enough for the approach to succeed, but neither low-level policy maintains force on a grasped object, so the box is frequently lost while being carried.

\subsection{Variation of Performance}

Policies generated independently for the same task differ widely in quality. Averaged over the five tasks and both low-level policies, a single generated policy (Best-of-$1$) reaches $43.9\%$ success with a standard deviation of $23.4$.
This variation arises mainly because a task usually admits several sensible task logics that do not perform equally. On \texttt{PBX}, for instance, one recurring logic drives the box along the line joining the humanoid, the box, and the target, while another aligns with one face of the box and switches faces whenever the box drifts, which avoids rotating the box but rarely completes the task within the episode limit.

This diversity is a direct consequence of providing no examples and no hints about how a task should be solved. While it costs some consistency across runs, it is also what lets the method discover solutions instead of reproducing prescribed ones, and we are free to keep whichever one performs best. Such diversity is valuable in robotics, where the strategy that works best is rarely known in advance and often depends on the environment setup.

\begin{figure*}[t]
    \centering
    \begin{minipage}[b]{0.63\textwidth}
        \centering
        \begin{subfigure}[b]{0.32\linewidth}
            \centering
            \includegraphics[width=0.9\linewidth]{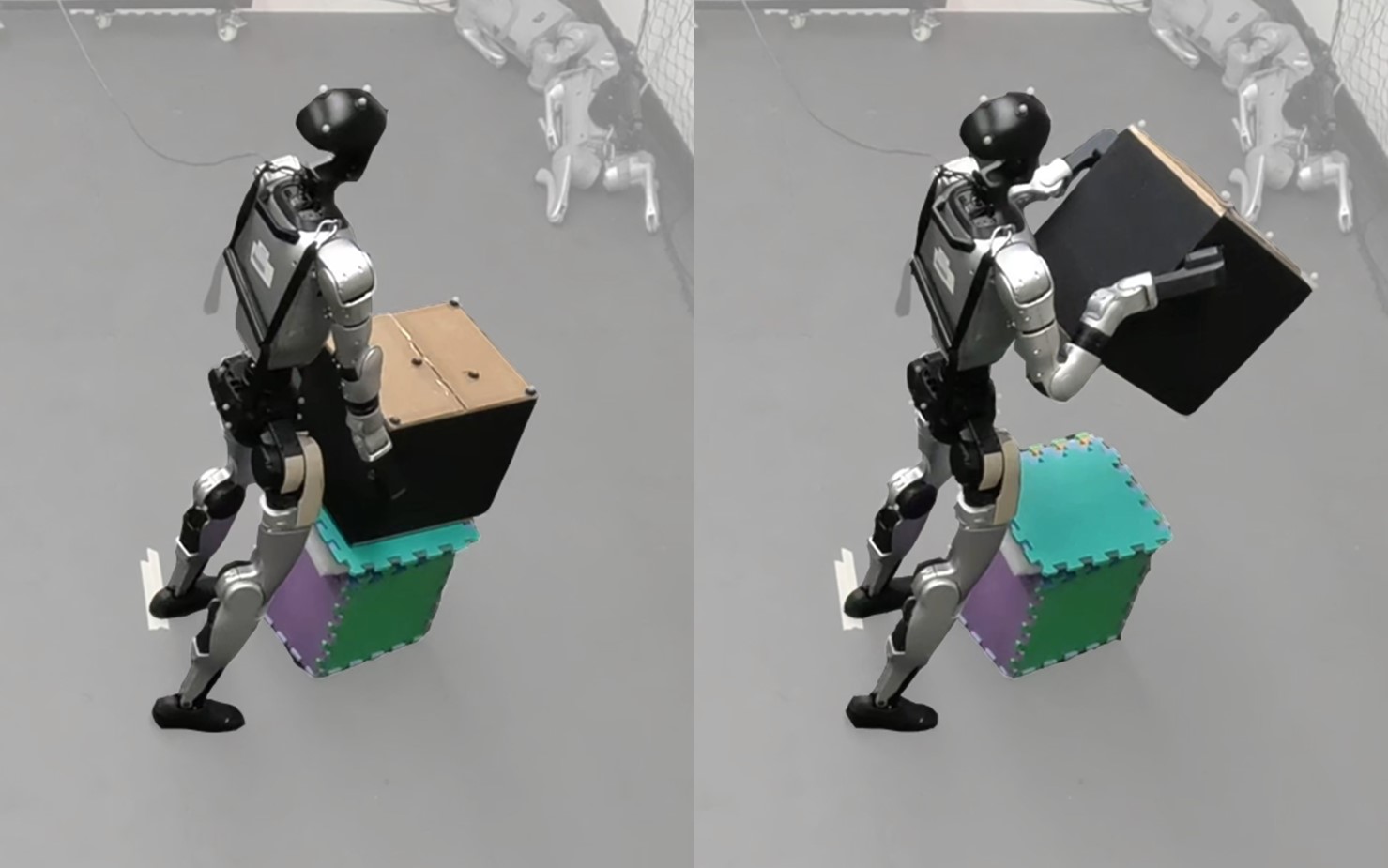}
            \caption{Height $0.4$m}
        \end{subfigure}
        \hfill
        \begin{subfigure}[b]{0.32\linewidth}
            \centering
            \includegraphics[width=0.9\linewidth]{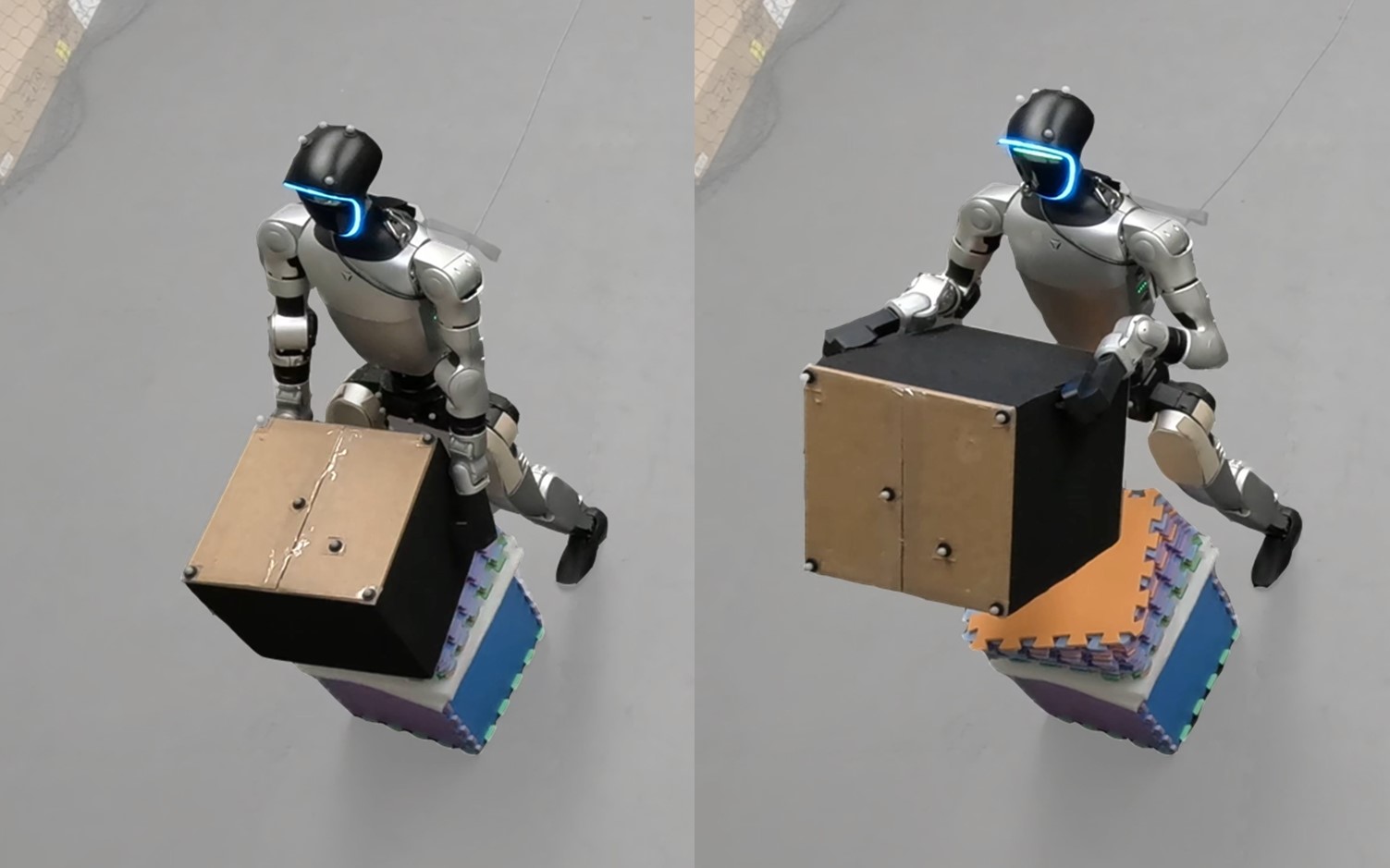}
            \caption{Height $0.5$m}
        \end{subfigure}
        \hfill
        \begin{subfigure}[b]{0.32\linewidth}
            \centering
            \includegraphics[width=0.9\linewidth]{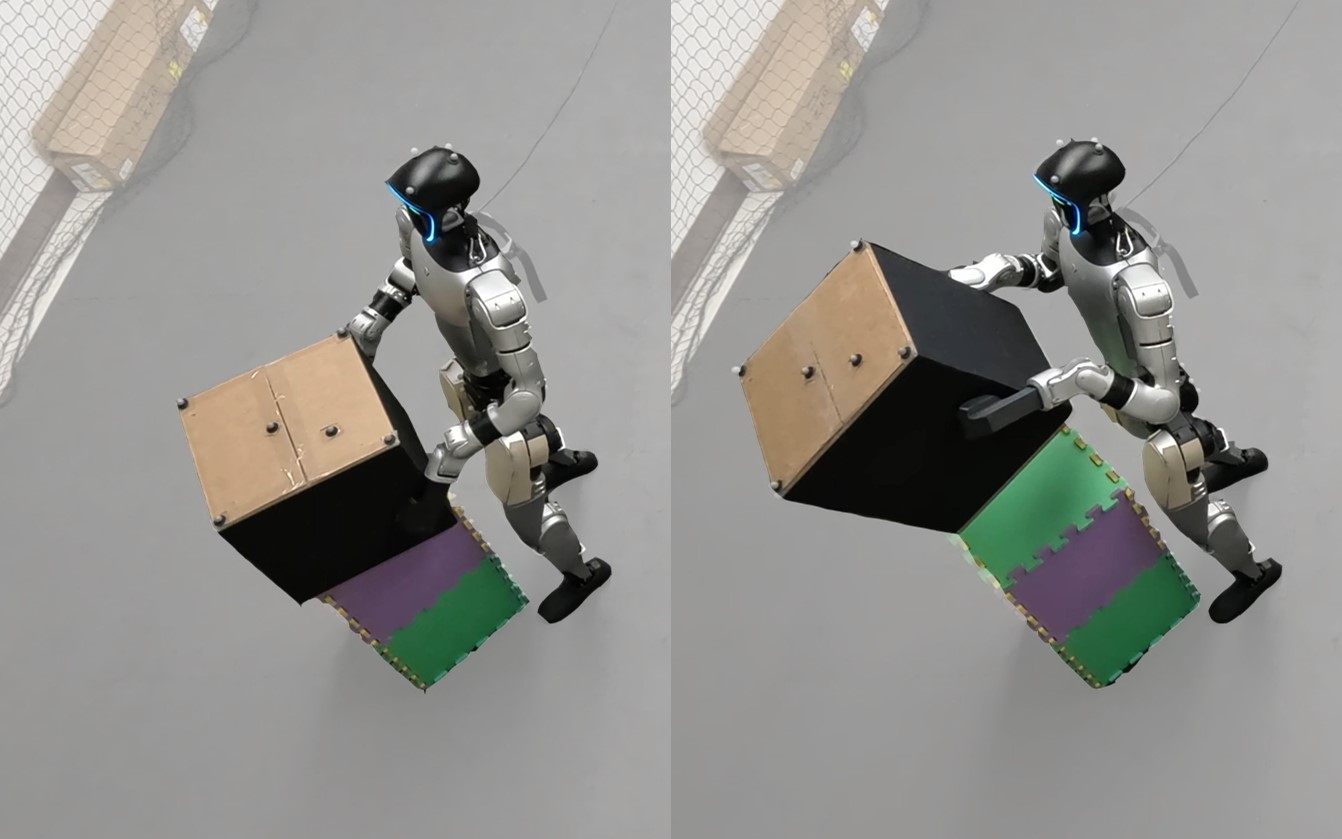}
            \caption{Height $0.62$m}
        \end{subfigure}
        \caption{\texttt{LBX} on hardware, all from the same policy code. The policy sets its body height and end-effector targets from the height of the box rather than replaying a fixed motion.}
        \label{fig:lbx_conditions}
    \end{minipage}
    \hfill
    \begin{minipage}[b]{0.34\textwidth}
        \centering
        \scriptsize
        \begin{tabular}{l|cc|cc}
            \toprule
             & \multicolumn{2}{c|}{\texttt{PBT}} & \multicolumn{2}{c}{\texttt{LBX}} \\
            Policy & Sim & Real & Sim & Real \\
            \midrule
            1 & $90$ & $93.3$ & $78$ & $66.7$ \\
            2 & $90$ & $20.0$ & $70$ & $20.0$ \\
            3 & $79$ & $73.3$ & $64$ & $6.7$ \\
            \bottomrule
        \end{tabular}
        \captionof{table}{Success rates (\%) of the three policy codes per task with the highest simulation success rate, evaluated in simulation ($100$ episodes) and on hardware ($15$ trials).}
        \label{tab:hardware_transfer}
    \end{minipage}
    \vspace{-9pt}
\end{figure*}

\subsection{Cost of Reliability} \label{subsec:cost}

The variation described above means that a single generated policy is unreliable, which is why we perform best-of-$N$ selection. Fig.~\ref{fig:cost} shows what this selection achieves against its API cost and its wall-clock time. Increasing $N$ both raises the expected success rate and reduces its variation. Averaged over the five tasks and both low-level policies, the mean success rate rises from $43.9\%$ to $79.1\%$ from $N=1$ to $N=10$, while its standard deviation falls from $23.4$ to $6.6$. Best-of-$N$ selection substantially improves reliability.

\textbf{API cost.} Generating and refining a single policy costs approximately \$$1.1$, so obtaining a best-of-$10$ policy costs approximately \$$11$. The curve increases monotonically but flattens beyond $N\approx8$, so we select $N=10$ as our operating point. Therefore, obtaining a policy for a new task costs on the order of ten dollars, which is negligible compared to the cost of designing rewards or collecting expert demonstrations.

\textbf{Wall-clock time.} We report wall-clock time separately for each low-level policy. Most of it is spent simulating rollouts rather than querying the LLM, and the two policies differ only because \texttt{sonic} plans one environment at a time while \texttt{rl\_isaac} can be evaluated as a batch. Obtaining a policy still takes a few hours, whereas the high-level RL baseline takes more than two days to train and needs a hand-designed reward for every task. Note that two days covers a single training run. Any change to the reward starts it over.

\subsection{Ablation Studies}

The quality of our refinement depends on what the LLM sees when it evaluates a rollout, namely the video frames and the numerical trajectory. To measure how each of these contributes, we select zero-shot policies with near-zero success rate for each task other than \texttt{LGP}, whose zero-shot policies already succeed, and repeat the refinement while varying only what the LLM receives. Table~\ref{tab:ablation_evaluation} reports the results.

\textbf{Is visual information necessary?} Since the numerical trajectory already contains the observations and internal states of the policy code, one may ask whether the video frames are needed at all. \textbf{Text Only} provides the numerical trajectory alone. Removing the frames cuts the success rate averaged over tasks by more than $40\%$, showing that visual information is necessary. Numbers alone do not reveal how the humanoid is positioned at the moment of contact with the object, which is precisely what distinguishes a policy that approaches the object correctly from one that does not.

\textbf{Does it matter which frames we show?} Having established that the visual frames are needed, we next ask how they should be chosen. \textbf{Even Only} uses evenly spaced frames throughout the episode, and \textbf{Keyframe Only} uses only the keyframes selected by the LLM without supplementing them with evenly spaced frames. Both perform worse than combining the two, confirming that the evaluation requires broad temporal coverage as well as the specific moments at which the task succeeds or fails.

\section{Hardware Experiments}

\subsection{Hardware Setup}

We deploy our high-level policies on a Unitree G1 humanoid using the same \texttt{sonic} low-level policy as in simulation. The observation $o_t$ is constructed from a motion capture system. The observation and command specifications match those used in simulation, thereby allowing the generated policy code to run on hardware without modification.

\subsection{Zero-Shot Transfer of Simulation-Generated Policies}
A policy that succeeds in simulation is meaningful if it also succeeds on the physical robot, where contact, friction, and command tracking all differ from the simulation. We therefore evaluate zero-shot transfer on \texttt{LBX} and \texttt{PBT}, the two tasks that fit within our hardware workspace. For \texttt{LBX}, we use policies generated with the hint in Sec.~\ref{subsec:main_results}. For each task, we deploy three policy codes with the highest simulation success rate and run $15$ trials per policy, using the same initial conditions for every policy. Table~\ref{tab:hardware_transfer} reports the results.

The best policy for each task transfers well zero-shot, succeeding in $93.3\%$ of trials on \texttt{PBT} and $66.7\%$ on \texttt{LBX}, close to its simulation success rate. Fig.~\ref{fig:pbt_conditions} and Fig.~\ref{fig:lbx_conditions} show the best-performing \texttt{PBT} and \texttt{LBX} policies running on the robot across different initial conditions, which each policy handles without any change between trials. However, zero-shot transfer is not equally successful for every policy. Some of the policies lose part of their success rate on hardware.

These losses are primarily associated with mismatches in low-level tracking and contact dynamics between simulation and hardware.
The low-level policy tracks small planar velocity commands even less reliably on the hardware than in simulation. On \texttt{LBX}, the contact-force limitation we observe in simulation becomes more pronounced, and the box slips out of the end-effectors more often during the lift.
A low-level policy better suited for contact-rich interaction could address this limitation without changing our high-level policy generation framework.

\subsection{Refinement on Hardware}

\begin{figure}[t]
    \centering
    \includegraphics[width=0.6\linewidth]{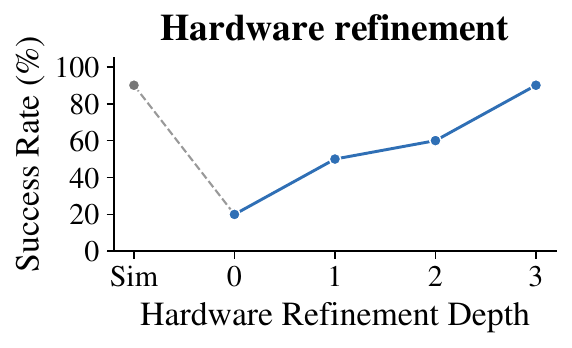}
    \caption{Success rate at each refinement depth on hardware, starting from the policy that transfers worst. The simulation success rate of the same policy is shown for reference.}
    \label{fig:hardware_depth}
\end{figure}

The zero-shot transfer results show that simulation-generated policies can transfer successfully to hardware, but may still suffer from sim-to-real discrepancies. Our refinement loop requires only rollouts of the current policy and can be applied directly on hardware without changing the prompts, evaluation, or refinement logic.

We demonstrate this on the \texttt{PBT} policy that transfers worst, which drops from $90\%$ in simulation to $20\%$ on hardware. 
Starting from this policy, we run the refinement loop on hardware, where the LLM analyzes the real-world rollouts to diagnose failure modes that emerge after transfer. Each candidate is evaluated over $10$ trials.
As shown in Fig.~\ref{fig:hardware_depth}, the success rate increases at every refinement depth, from $20\%$ to $50\%$, $60\%$, and finally $90\%$, matching the success rate the policy achieved in simulation. For this policy, three refinement depths, which required four rounds of refinement in total, recover its simulation performance.

\begin{figure}[t]
    \centering
    \includegraphics[width=1.0\linewidth]{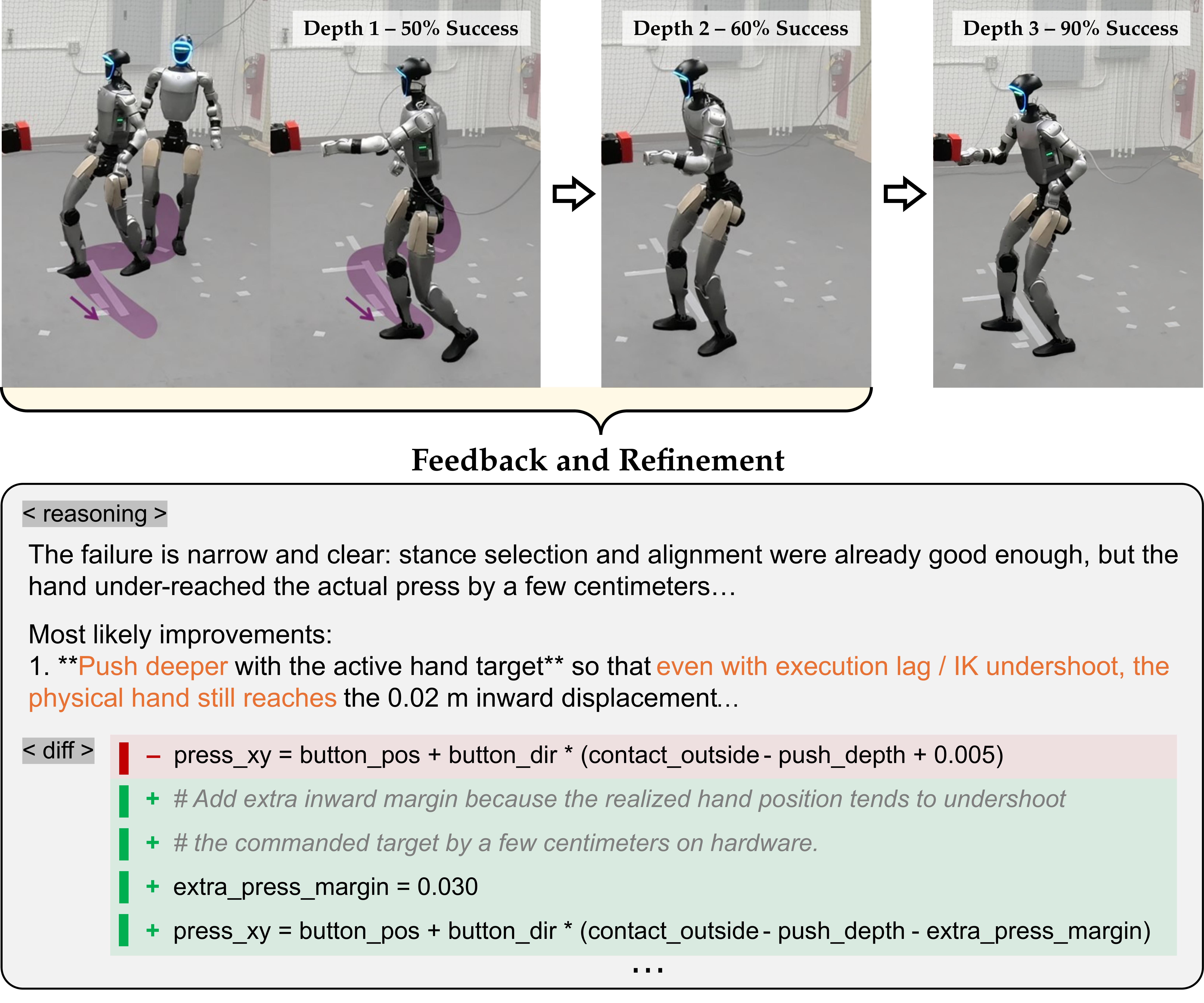}
    \caption{Example of one refinement depth on hardware, generated by the LLM without any human feedback. Top: executions at refinement depths $1$ to $3$. Bottom: at depth $1$, the rollout feedback attributes the failure to the end-effector under-reaching the button, and the resulting code change introduces an explicit inward margin to compensate for the undershoot.}
    \label{fig:hardware_refinement}
\end{figure}

The refinement identifies a hardware-specific failure mode: navigation succeeds, but the end-effector under-reaches the button.
Fig.~\ref{fig:hardware_refinement} shows the refinement from depth $1$ to depth $2$, where the feedback identifies a few-centimeter undershoot and adds an inward margin to the press target. At depth $3$, the evaluation reports that the end-effector still falls short in some trials, and the refinement enlarges the same margin and holds the press longer, reaching $90\%$. 
The simulation does not exhibit this undershoot, so refinement in simulation would not have produced this correction.

This recovery is achieved entirely from hardware rollouts, using the same refinement pipeline as in simulation and without real-world expert demonstrations or policy retraining. This demonstrates that HuGo can correct sim-to-real discrepancies through refinement from hardware rollouts.


\section{Conclusion}

We introduce HuGo, which generates closed-loop high-level policy code for humanoid loco-manipulation on top of a frozen whole-body policy, without expert demonstrations, reference motions, or reward design. 
The LLM generates the policy code and refines it using feedback from the policy's rollouts. Across five tasks and two low-level policies, the generated policies adapt to randomized initial conditions, and support zero-shot transfer to hardware. We further demonstrate that the same refinement loop can correct sim-to-real discrepancies from hardware rollouts, raising a \texttt{Push Button} policy from $20\%$ to $90\%$ success.

Our method inherits the limits of the low-level policy and the LLM. A low-level policy that maintains contact force, or one with a dexterous hand, would raise success on tasks that require holding an object. The LLM sets the second ceiling, since the policy code can express a collision-free approach that the LLM does not infer on its own. For future work, a cross-task memory would let lessons from earlier tasks improve the code the LLM writes for a new one. The generated policies could also supervise a vision-based policy by distillation, or seed an RL replay buffer.

\section*{ACKNOWLEDGMENT}

The authors thank Lasse Peters and Hongrui Zhao for their feedback on the manuscript, and Joohwan Seo for his feedback and for the end-effector design used on the Unitree G1.



\bibliographystyle{ieeetr}
\bibliography{bibliography}

@inproceedings{ma2024eureka,
  title={Eureka: Human-level reward design via coding large language models},
  author={Ma, Yecheng Jason and others},
  booktitle={International conference on learning Representations},
  volume={2024},
  pages={26516--26560},
  year={2024}
}

@inproceedings{ryu2025curricullm,
  title={Curricullm: Automatic task curricula design for learning complex robot skills using large language models},
  author={Ryu, Kanghyun and others},
  booktitle={2025 IEEE International Conference on Robotics and Automation (ICRA)},
  pages={4470--4477},
  year={2025},
  organization={IEEE}
}

@article{choi2025craft,
  title={Craft: Coaching reinforcement learning autonomously using foundation models for multi-robot coordination tasks},
  author={Choi, Seoyeon and others},
  journal={arXiv preprint arXiv:2509.14380},
  year={2025}
}

@article{romera2024mathematical,
  title={Mathematical discoveries from program search with large language models},
  author={Romera-Paredes, Bernardino and others},
  journal={Nature},
  volume={625},
  number={7995},
  pages={468--475},
  year={2024},
  publisher={Nature Publishing Group UK London}
}

@article{ye2024reevo,
  title={Reevo: Large language models as hyper-heuristics with reflective evolution},
  author={Ye, Haoran and others},
  journal={Advances in neural information processing systems},
  volume={37},
  pages={43571--43608},
  year={2024}
}

@article{novikov2025alphaevolve,
  title={Alphaevolve: A coding agent for scientific and algorithmic discovery},
  author={Novikov, Alexander and others},
  journal={arXiv preprint arXiv:2506.13131},
  year={2025}
}

@misc{karpathy2026autoresearch,
  author       = {Andrej Karpathy},
  title        = {AutoResearch},
  year         = {2026},
  howpublished = {\url{https://github.com/karpathy/autoresearch}},
  note         = {GitHub repository}
}

@article{xie2026grail,
  title={GRAIL: Generating Humanoid Loco-Manipulation from 3D Assets and Video Priors},
  author={Xie, Tianyi and others},
  journal={arXiv preprint arXiv:2606.05160},
  year={2026}
}

@article{weng2025hdmi,
  title={Hdmi: Learning interactive humanoid whole-body control from human videos},
  author={Weng, Haoyang and others},
  journal={arXiv preprint arXiv:2509.16757},
  year={2025}
}

@article{schuck2026learning,
  title={Learning Loco-Manipulation From SMPC Demonstrations With Sparse Offline-to-Online RL},
  author={Schuck, Martin and others},
  journal={arXiv preprint arXiv:2608.12063},
  year={2026}
}

@article{taouil2026motiondisco,
  title={MotionDisco: Motion Discovery for Extreme Humanoid Loco-Manipulation},
  author={Taouil, Ilyass and others},
  journal={arXiv preprint arXiv:2606.06139},
  year={2026}
}

@article{liu2025ego,
  title={Ego-vision world model for humanoid contact planning},
  author={Liu, Hang and others},
  journal={arXiv preprint arXiv:2510.11682},
  year={2025}
}

@article{luo2026sonic,
  title={Sonic: Supersizing motion tracking for natural humanoid whole-body control},
  author={Luo, Zhengyi and others},
  journal={Science Robotics},
  volume={11},
  number={117},
  pages={eaed4592},
  year={2026},
  publisher={American Association for the Advancement of Science}
}

@article{schulman2017proximal,
  title={Proximal policy optimization algorithms},
  author={Schulman, John and others},
  journal={arXiv preprint arXiv:1707.06347},
  year={2017}
}

@inproceedings{liang2023code,
  title={Code as policies: Language model programs for embodied control},
  author={Liang, Jacky and others},
  booktitle={2023 IEEE International Conference on Robotics and Automation (ICRA)},
  pages={9493--9500},
  year={2023},
  organization={IEEE}
}

@article{fu2026capx,
  title={CaP-X: A framework for benchmarking and improving coding agents for robot manipulation},
  author={Fu, Letian and others},
  journal={arXiv preprint arXiv:2603.22435},
  year={2026}
}

@article{chen2026gap,
  title={GaP: A graph-as-policy multi-agent self-learning harness for variational automation tasks},
  author={Chen, Kaiyuan and others},
  journal={arXiv preprint arXiv:2607.05369},
  year={2026}
}

@inproceedings{wang2024autonomous,
  title={Autonomous behavior planning for humanoid loco-manipulation through grounded language model},
  author={Wang, Jin and others},
  booktitle={2024 IEEE/RSJ International Conference on Intelligent Robots and Systems (IROS)},
  year={2024},
  organization={IEEE}
}

@article{wen2025humanoidcoa,
  title={Humanoid agent via embodied chain-of-action reasoning with multimodal foundation models for zero-shot loco-manipulation},
  author={Wen, Congcong and others},
  journal={arXiv preprint arXiv:2504.09532},
  year={2025}
}

@article{sygkounas2026memento,
  title={MEMENTO: Memory-guided memetic code-as-policy evolution},
  author={Sygkounas, Alkis and others},
  journal={arXiv preprint arXiv:2607.22832},
  year={2026}
}

@article{du2023vlmsuccess,
  title={Vision-language models as success detectors},
  author={Du, Yuqing and others},
  journal={arXiv preprint arXiv:2303.07280},
  year={2023}
}

@article{zhao2025resmimic,
  title={ResMimic: From general motion tracking to humanoid whole-body loco-manipulation via residual learning},
  author={Zhao, Siheng and others},
  journal={arXiv preprint arXiv:2510.05070},
  year={2025}
}

@article{li2026mpcrl,
  title={Accelerating and scaling MPC-guided reinforcement learning for humanoid locomotion and manipulation},
  author={Li, Junheng and others},
  journal={arXiv preprint arXiv:2606.05687},
  year={2026}
}

@article{kuang2025skillblender,
  title={SkillBlender: Towards versatile humanoid whole-body loco-manipulation via skill blending},
  author={Kuang, Yuxuan and others},
  journal={arXiv preprint arXiv:2506.09366},
  year={2025}
}

@article{wang2023prompt,
  title={Prompt a robot to walk with large language models},
  author={Wang, Yen-Jen and others},
  journal={arXiv preprint arXiv:2309.09969},
  year={2023}
}

@misc{anthropic2026claude,
  title={How {Claude} performs on robotics tasks},
  author={{Anthropic}},
  year={2026},
  howpublished={\url{https://www.anthropic.com/research/claude-plays-robotics}}
}

@article{hosseinzadeh2026kite,
  title={KITE: Keyframe-indexed tokenized evidence for VLM-based robot failure analysis},
  author={Hosseinzadeh, Mehdi and others},
  journal={arXiv preprint arXiv:2604.07034},
  year={2026}
}

@article{yang2026handoff,
  title={HANDOFF: Humanoid agentic task-space whole-body control via distilled complementary teachers},
  author={Yang, Lizhi and others},
  journal={arXiv preprint arXiv:2606.06493},
  year={2026}
}

@inproceedings{seo2023deep,
  title={Deep imitation learning for humanoid loco-manipulation through human teleoperation},
  author={Seo, Mingyo and others},
  booktitle={2023 IEEE-RAS 22nd International Conference on Humanoid Robots (Humanoids)},
  pages={1--8},
  year={2023},
  organization={IEEE}
}

@article{bohez2022imitate,
  title={Imitate and repurpose: Learning reusable robot movement skills from human and animal behaviors},
  author={Bohez, Steven and others},
  journal={arXiv preprint arXiv:2203.17138},
  year={2022}
}

@article{liao2025beyondmimic,
  title={BeyondMimic: From motion tracking to versatile humanoid control via guided diffusion},
  author={Liao, Qiayuan and others},
  journal={arXiv preprint arXiv:2508.08241},
  year={2025}
}

@inproceedings{radosavovic2024terrain,
  title={Learning humanoid locomotion over challenging terrain},
  author={Radosavovic, Ilija and others},
  booktitle={Conference on Robot Learning (CoRL)},
  year={2024}
}


\end{document}